\documentclass{article}
\usepackage{amsmath,mathtools,mathptmx} 
\usepackage{tikz, tikzscale, pgf, pgfplots, pgflibraryarrows}
\usepackage{natbib}
\usetikzlibrary{shapes.geometric}
\usepackage{url}
\usepackage{hyperref}
\usepackage{algorithm}
\usepackage{algorithmic}
\usetikzlibrary{positioning}
\usepackage{caption, subcaption}
\usepackage{comment}
\usepackage[accepted]{icml2016}

\icmltitlerunning{Efficient Parallel Learning of Word2Vec}

\pgfdeclarelayer{back}
\pgfsetlayers{back,main}

\begin{document} 

\twocolumn[
\icmltitle{Efficient Parallel Learning of Word2Vec}

\icmlauthor{Jeroen B.P. Vuurens}{j.b.p.vuurens@tudelft.nl}
\icmladdress{The Hague University of Applied Science\\
		    Delft University of Technology, The Netherlands}
\icmlauthor{Carsten Eickhoff}{carsten.eickhoff@inf.ethz.ch}
\icmladdress{ETH Zurich\\
Department of Computer Science, Zurich, Switzerland}
\icmlauthor{Arjen P. de Vries}{arjen@acm.org}
\icmladdress{Radboud University Nijmegen\\
		Institute for Computing and Information Sciences, Nijmegen, The Netherlands}

\icmlkeywords{deep learning, parallel processing, efficiency}
\vskip 0.2in
]

\begin{abstract}
Since its introduction, Word2Vec and its variants are widely used to learn semantics-preserving representations of words or entities in an embedding space, which can be used to produce state-of-art results for various Natural Language Processing tasks. Existing implementations aim to learn efficiently by running multiple threads in parallel while operating on a single model in shared memory, ignoring incidental memory update collisions. We show that these collisions can degrade the efficiency of parallel learning, and propose a straightforward caching strategy that improves the efficiency by a factor of 4.
\end{abstract}


\section{Introduction}

Traditional NLP approaches dominantly use simple bag-of-word representations of documents and sentences, but recent approaches that use distributed representation of words, by constructing a so-called "word embedding", have been shown effective across many different NLP tasks \cite{weston2014tagspace}. Mikolov et al. introduced efficient strategies to learn embeddings from text corpora, which are collectively known as Word2Vec \cite{mikolov2013distributed}. They have shown that simply increasing the volume of training data improves the semantic and syntactic generalizations that are automatically encoded in embedding space, and therefore efficiency is the key to increase the potential of this technique.

In recent work, Ji et al. argue that the current implementations of Word2Vec do not scale well over the number of used cores, and propose a solution that uses higher-level linear algebra functions to improve the efficiency of Word2Vec using negative sampling \cite{ji2016parallelizing}. In this work, we analyze the scalability problem and conclude it is mainly caused by simultaneous access of the same memory by multiple threads. As a straightforward solution, we propose to cache frequently updated vectors, and show a large gain in efficiency as a result.


\section{Related Work}

Bengio et al.\ propose to learn a distributed representation for words in an embedding space based on the words that immediately precede a word in natural language \cite{bengio2006neural}.  They use a shallow neural network architecture to learn a vector-space representation for words comparable to those obtained in latent semantic indexing. Recently, Mikolov et al.\ proposed a very efficient way to learn embeddings over large text corpora called Word2Vec \cite{mikolov2013distributed}, which produces state-of-the-art results on various tasks, such as word similarity, word analogy tasks, named entity recognition, machine translation and question answering \cite{weston2014tagspace}. Word2vec uses Stochastic Gradient Descent to optimize the vector representations by iteratively learning on words that appear jointly in text.

For efficient learning of word embeddings, Mikolov et al.\ proposed two strategies that avoid having to update the probability of observing every word in the vocabulary. \textbf{Negative sampling} consists of updating observed word pairs together with a limited number of random word pairs to regularize weights. Alternatively, a \textbf{hierarchical softmax} is used as a learning objective, for which the output layer is transformed into a binary Huffmann tree and instead of predicting an observed word, the model learns the word's position in the Huffmann tree by the left and right turns at each inner node along its path from the root.

The original Word2Vec implementation uses a Hogwild strategy of allowing processors lock-free access to shared memory, in which they can update at will \cite{recht2011hogwild}. Recht et al.\ show that when data access is sparse memory overwrites are rare and barely introduce error when they occur. However, Ji et al.\ notice that for training Word2Vec this strategy does not use computational resources efficiently \cite{ji2016parallelizing}. They propose to use level-3 BLAS routines on specific CPU's to operate on a mini-batches, which is then updated to the main memory and reduces the access to main memory. They address the Word2Vec variant that uses negative sampling. In our work we address the hierarchical softmax variant, and show that straightforward caching of frequently occurring data is sufficient to improve efficiency on conventional hardware.

\section{Experiment}
\label{section:experiment}

\subsection{Analysis of inefficiencies for learning Word2Vec}

For faster training of models, commonly, the input is processed in parallel by multiple processors. One strategy is to map the input into separate batches and aggregating the results in a reduce step. Alternatively, multithreading can be used on a single model in shared memory, which is used by the original Word2Vec implementation in C as well as the widely used Gensim package for Python. We analyzed how the learning times scale over the used number of cores with Word2Vec, and noticed the total run time appears to be optimal when using approximately 8 cores, beyond which efficiency degrades (see Figure~\ref{fig:speed}). 

Ji et al. analyze this problem and conclude that multiple threads can cause conflicts when they attempt to update the same memory \cite{ji2016parallelizing}. Current Word2Vec implementations simply ignore memory conflicts between threads, however, concurrent attempts to read and update the same memory increases memory access latency. This Hogwild strategy was introduced under the assumption that memory overwrites are rare, however, for natural language memory collisions may be more likely than for other data given the skew towards frequently occurring words. Collisions are even more likely when learning against a hierarchical softmax, since the top nodes in the Huffmann tree are used by a large part of the vocabulary. 

\subsection{Caching frequently used vectors}

For efficient learning of word embeddings with negative sampling, Ji et al. propose a mini-batch strategy that uses a level-3 BLAS function to multiply a matrix that consists of all words that share the same context word (e.g. with a window size of 5 up to 10 words would train against the same context term), with a matrix that contains the context word and a shared set of negative samples \cite{ji2016parallelizing}. Efficiency is improved up to 3.6x over the original Word2Vec implementation, by the leveraging the improved ability for matrix-matrix multiplication in high-end CPU's, and only updating the model after each mini-batch. Typically, the batches are limited to about 10-20 words for convergence reasons, reporting only a marginal loss of accuracy. 

We suspect that for parallel training a model in shared memory the real bottleneck is latency caused by concurrent access of the same memory. The solution that is presented by \cite{ji2016parallelizing} uses level-3 BLAS functions to perform matrix-matrix multiplications, which implicitly lowers the number of times shared memory is accessed and updated. We alternatively propose a straightforward caching strategy that provides a comparable efficiency gain over the hierarchical softmax variant, that is less hardware dependent since it does not need highly optimized level-3 BLAS functions. When using the hierarchical softmax, efficiency is more affected by memory conflicts than with negative sampling; the root node of the tree is in every word's path and therefore gives a conflict every time, the direct children of the root will be in the path of half the vocabulary, etc. When each thread caches the most frequently used upper nodes of the tree and only updates the change to main memory after a number of trained words, the number of memory conflicts is reduced.

\begin{algorithm}[t]
\caption{Cached Skipgram HS}
\label{cythnn}
\begin{algorithmic}[1]
\FOR{each wordid $v$ in text}
\FOR{each wordid $w$ within window of $v$}
\FOR{each node $n$, turn $t$ in treepath to $v$}
\IF{isFrequent($n$)}
\IF{not cached[$n$]}
\STATE scopy( weight[$n$], cachedweight[ $n$ ] )
\STATE scopy( cachedweight[ $n$ ], original[ $n$ ] )
\STATE cached[$n$] = True
\ENDIF
\STATE usedweight=cachedweight[ $n$ ]
\ELSE
\STATE usedweight=weight[ $n$ ]
\ENDIF
\STATE f = sigmoid( sdot( vector[ $w$], usedweight ) )
\STATE gradient = ( 1 - $t$ - f ) * alpha
\STATE saxpy( 1, usedweight, hiddenlayer )
\STATE saxpy( 1, vector[ $w$ ], usedweight )
\ENDFOR
\STATE saxpy( 1, hiddenlayer, vector[ $w$ ] )
\ENDFOR
\IF{time to flush cache}
\FOR{each node $n$}
\IF{isFrequent( $n$ ) and cached[ $n$ ]}
\STATE saxpy( -1, original[ $n$],  cachedweight[ $n$ ] )
\STATE saxpy( 1, cachedweight[ $n$ ], weight[ $n$ ] )
\STATE cached[ $n$ ] = False
\ENDIF
\ENDFOR
\ENDIF
\IF{time to update alpha}
\STATE update alpha
\ENDIF
\ENDFOR
\end{algorithmic}
\end{algorithm}

Algorithm~\ref{cythnn} describes the caching in pseudocode, in which for the most frequently used inner nodes in the Huffmann tree a local copy is created and used for the updates and the non frequently used inner nodes are updated in shared memory. After processing a number of words (typically in the range 10 to 100) for every cached weight vector the update is computed by subtracting the original value when cached and adding this update to the weight vector in main memory. The vectors are processed using level-1 BLAS functions, \texttt{scopy} to copy a vector, \texttt{sdot} to compute the dot product between two vectors and \texttt{saxpy} to add a scalar multiplied by the first vector to the second vector.

The proposed caching strategy is implemented in Python/Cython and published as part of the Cythnn open source project\footnote{\url{http://cythnn.github.io}}. 


\section{Results}
\label{section:results}

\subsection{Experiment setup}

To compare the efficiency of the proposed caching strategy with the existing implementations, we used the skipgram with hierarchical softmax, with the default settings mintf=5, windowsize=5, vectorsize=100, downsample=0.001, and without negative sampling. Although our implementation is only slightly improved by downsampling of frequent terms, the original C implementation is approximately 24\% faster and produces slightly more accurate word embeddings on the analogy test. Every run processed 10 iterations over the standard Text8 collection\footnote{\url{http://mattmahoney.net/dc/text8.zip }}, which consists of the first 100MB of flat text words found in Wikipedia. All experiments were performed on the same virtual machine with two Intel(R) Xeon(R) CPU E5-2698 v3, which together have 32 physical cores.

\subsection{Efficiency}

\begin{figure}[th]
   \centering
       \includegraphics[]{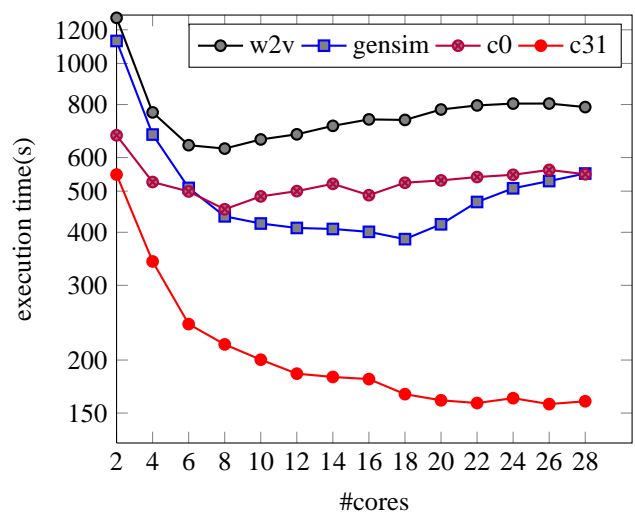}
       \caption{Comparison of the total execution time when changing the number of cores (x-axis) between the original Word2Vec, Gensim, Cythnn without caching and Cythnn when caching the top-31 nodes in the Huffmann tree}
    \label{fig:speed}
\end{figure}

In Figure~\ref{fig:speed}, we compare the changes in total run time when adding more cores between the original Word2Vec C implementation, Gensim, Cythnn without caching (c0) and Cythnn when caching the 31 most frequently used nodes in the Huffmann tree (c31). The first observation is that Word2Vec and Cythnn without caching do have degrading efficiencies beyond the use of 8 cores, while Gensim is apparently more optimized to use up to 16 cores. The run of Cythnn that caches the 31 top-vectors (157 sec) performs up to 3.9x faster than the fastest C-run (630 sec) and up to 2.5x faster than the fastest Gensim run (385 sec).

\begin{figure}[t]
   \centering
       \includegraphics[]{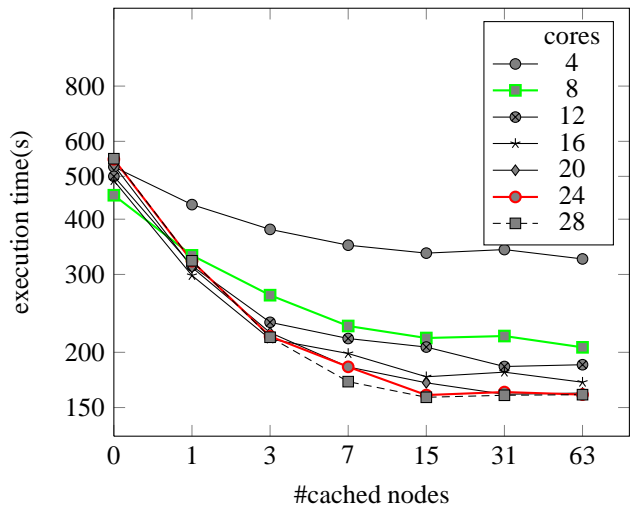}
       \caption{Comparison of the execution time when changing the number of cores and on the x-axis the number of most frequent terms from the Huffmann tree that are cached..}
    \label{fig:speed_cores}
\end{figure}

In Figure~\ref{fig:speed_cores} we compare the execution time of Cythnn when changing the number of cached top level nodes in the Huffmann tree over different numbers of cores used. Interestingly, caching just the root node (\#cached nodes=1) has the highest impact on efficiency, and reduces the execution time by more than 40\% when using more than 20 cores. This supports the idea that the reduction of concurrent access of the same memory is key to improving efficiency. Since the root node in the Huffmann tree is in the path of every word in the vocabulary, caching just this node improves most. On the Text8 dataset the improvement is near optimal when caching at least the top-15 nodes in the tree. In our experiments, using more than 24 cores is slightly counter effective, possibly indicating that memory bandwidth is fully used.

\subsection{Effectiveness}

Ji et al.\ compared the effectiveness between their approach and the original Word2Vec implementation and report that processing in mini-batches has a marginal but noticeable negative effect on the accuracy of the embeddings learned. In Table~\ref{table:accuracy}, we show the average accuracy of each system over the runs of 2-28 cores, where the accuracy was computed using the evaluation tool and question-words test set that is packaged with Word2Vec\footnote{\url{http://word2vec.googlecode.com}}. The results show a higher accuracy for Gensim, but we have found no documentation that helps to explain this difference. Using Cythnn, caching does not result in lower effectiveness as reported by \cite{ji2016parallelizing}, a difference being that our caching approach always uses updated vectors within a thread whereas the BLAS-3 solution does not use updated vectors within a mini-batch.

When varying cache update frequency u, we expect a trade-off between training efficiency and model accuracy. In practice, our experiments show, that this balance is a slowly shifting one that is further influenced by the chosen window size n. Using the default window size of n = 5, updating the cache after each training word (u = 1) is as efficient as flushing after u = 10 words. For this paper, cached updates were written back to shared memory every 10 words.

\begin{table}[t]
\centering
\caption{Comparison of the accuracy between systems}
\label{table:accuracy}
\begin{tabular}{|lr|}
\hline
System               & Accuracy  \\
\hline
W2V C        		& 33.73   \\
Gensim       		& 35.45   \\
Cythnn no cache 	& 33.54  \\
Cythnn cache top-31 & 33.57 \\
\hline
\end{tabular}
\end{table}


\section{Conclusion}

In this study, we analyzed the scalability of parallel learning of a single model in shared memory, most specifically learning Word2Vec using a Skipgram architecture against a hierarchical softmax. When different processes access and update the same memory, either because a skew in the data or because of the increased number of concurrent cores used, there is only limited benefit to adding more cores. The original Word2Vec implementation is most efficient when trained using 8 cores and becomes less efficient when more cores are added. We propose a straightforward caching strategy that caches the weight vectors that are used most frequently, and updates their change to main memory after a short delay reducing concurrent access of shared memory. We compared the efficiency of this strategy against existing Word2Vec implementations and show up to 4x improvement in efficiency.

\section*{Acknowledgment}

This work was carried out on the Dutch national e\-infrastructure with the support of SURF Foundation.

\bibliography{bibliography}  

\begin{thebibliography}{5}
\providecommand{\natexlab}[1]{#1}
\providecommand{\url}[1]{\texttt{#1}}
\expandafter\ifx\csname urlstyle\endcsname\relax
  \providecommand{\doi}[1]{doi: #1}\else
  \providecommand{\doi}{doi: \begingroup \urlstyle{rm}\Url}\fi

\bibitem[Bengio et~al.(2006)Bengio, Schwenk, Sen{\'e}cal, Morin, and
  Gauvain]{bengio2006neural}
Bengio, Yoshua, Schwenk, Holger, Sen{\'e}cal, Jean-S{\'e}bastien, Morin,
  Fr{\'e}deric, and Gauvain, Jean-Luc.
\newblock Neural probabilistic language models.
\newblock In \emph{Innovations in Machine Learning}, pp.\  137--186. Springer,
  2006.

\bibitem[Ji et~al.(2016)Ji, Satish, Li, and Dubey]{ji2016parallelizing}
Ji, Shihao, Satish, Nadathur, Li, Sheng, and Dubey, Pradeep.
\newblock Parallelizing word2vec in shared and distributed memory.
\newblock \emph{preprint arXiv:1604.04661}, 2016.

\bibitem[Mikolov et~al.(2013)Mikolov, Sutskever, Chen, Corrado, and
  Dean]{mikolov2013distributed}
Mikolov, Tomas, Sutskever, Ilya, Chen, Kai, Corrado, Greg~S, and Dean, Jeff.
\newblock Distributed representations of words and phrases and their
  compositionality.
\newblock In \emph{Advances in neural information processing systems}, pp.\
  3111--3119, 2013.

\bibitem[Recht et~al.(2011)Recht, Re, Wright, and Niu]{recht2011hogwild}
Recht, Benjamin, Re, Christopher, Wright, Stephen, and Niu, Feng.
\newblock Hogwild: A lock-free approach to parallelizing stochastic gradient
  descent.
\newblock In \emph{Advances in Neural Information Processing Systems}, pp.\
  693--701, 2011.

\bibitem[Weston et~al.(2014)Weston, Chopra, and Adams]{weston2014tagspace}
Weston, Jason, Chopra, Sumit, and Adams, Keith.
\newblock \# tagspace: Semantic embeddings from hashtags.
\newblock In \emph{Proceedings of EMNLP 2014}, 2014.

\end{thebibliography}
\bibliographystyle{icml2016}

\end{document}